\documentclass[sigconf]{acmart}

\usepackage{multirow}

\AtBeginDocument{%
  \providecommand\BibTeX{{%
    \normalfont B\kern-0.5em{\scshape i\kern-0.25em b}\kern-0.8em\TeX}}}

\copyrightyear{2021}
\acmYear{2021}
\setcopyright{acmcopyright}\acmConference[MM '21]{Proceedings of the 29th ACM International Conference on Multimedia}{October 20--24, 2021}{Virtual Event, China}
\acmBooktitle{Proceedings of the 29th ACM International Conference on Multimedia (MM '21), October 20--24, 2021, Virtual Event, China}
\acmPrice{15.00}
\acmDOI{10.1145/3474085.3479218}
\acmISBN{978-1-4503-8651-7/21/10}

\begin{document}
\fancyhead{}

\title{Multi-modal Representation Learning for Video Advertisement Content Structuring}


\author{Daya Guo}
\authornote{Equal contribution. Order determined by alphabeta order.}
\affiliation{%
\institution{Sun Yat-sen University}
\city{Guangzhou}
\state{Guangdong}
\country{China}}
\email{guody5@mail2.sysu.edu.cn}

\author{Zhaoyang Zeng*}
\affiliation{%
\institution{Sun Yat-sen University}
\city{Guangzhou}
\state{Guangdong}
\country{China}}
\email{zengzhy5@mail2.sysu.edu.cn}

\begin{abstract}
Video advertisement content structuring aims to segment a given video advertisement and label each segment on various dimensions, such as presentation form, scene, and style. 
Different from real-life videos, video advertisements contain sufficient and useful multi-modal content like caption and speech, which provides crucial video semantics and would enhance the structuring process. 
In this paper, we propose a multi-modal encoder to learn multi-modal representation from video advertisements by interacting between video-audio and text. 
Based on multi-modal representation, we then apply Boundary-Matching Network to generate temporal proposals. To make the proposals more accurate, we refine generated proposals by scene-guided alignment and re-ranking. 
Finally, we incorporate proposal located embeddings into the introduced multi-modal encoder to capture temporal relationships between local features of each proposal and global features of the whole video for classification.
Experimental results show that our method achieves significantly improvement compared with several baselines and Rank 1 on the task of Multi-modal Ads Video Understanding in ACM Multimedia 2021 Grand Challenge. Ablation study further shows that leveraging multi-modal content like caption and speech in video advertisements  significantly improve the performance.
\end{abstract}




\maketitle

\section{Introduction}
\begin{figure}[h]
	%\vskip 0.2in
	\begin{center}
		\includegraphics[width=0.47\textwidth]{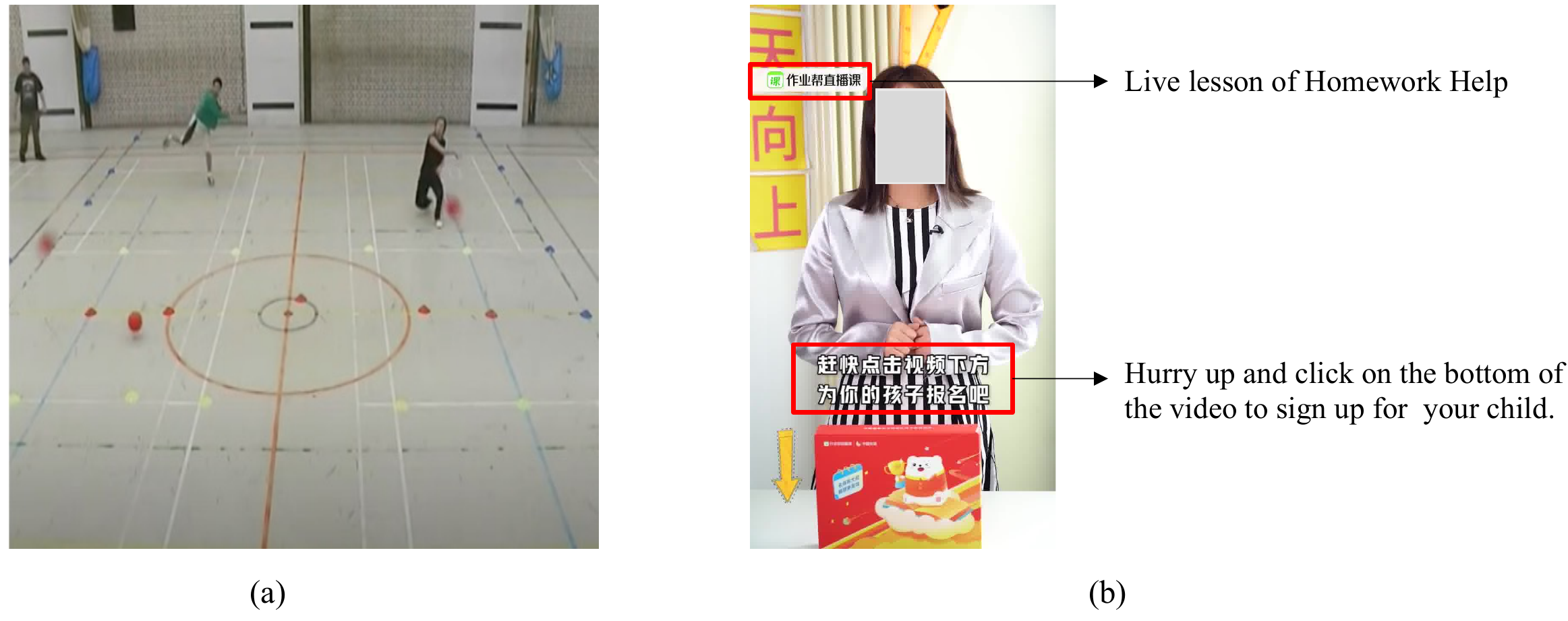}
		\caption{(a) The left part is a frame of video from ActivityNet dataset. (b) The right part is a frame of video advertisement from ACM Multimedia 2021 Grand Challenge.}
		\label{figure-case}
	\end{center}
\end{figure}

As the number of video advertisements in Internet grows rapidly, video ads content analysis methods have become more crucial and attracted more attention from both academia and industry. Video ads content structuring is an important task in video ads content analysis area, which aims to segment a given video ads in time and label each segment on various dimensions, such as presentation form, scene, and style. As shown in Figure \ref{figure-case}, video advertisements  are different from real-life videos in temporal action detection datasets like ActivityNet \cite{caba2015activitynet} and THUMOS \cite{idrees2017thumos}. They  provide sufficient multi-modal content like caption and speech for the purpose of promoting and popularizing their products. Therefore, multi-modal contents like caption and speech provide crucial video semantic and are important for the structuring process \cite{guo2019multi}.

In this paper, we propose a multi-modal encoder to learn multi-modal representation of video ads. The encoder consists of three components, including a video-audio encoder, a text encoder and a cross-modality encoder. The video-audio encoder contains several Inception modules \cite{szegedy2015going} implemented by 1D convlutional layers, which takes video and audio features as the input and outputs context representation of video-audio. The text encoder is a powerful pre-trained model BERT \cite{devlin2018bert}. A Transformer based cross-modality encoder is used for cross-modality interaction between the text and the video-audio to obtain multi-modal representation, which will be used for video advertisement content structuring.

Inspired by state-of-the-art methods \cite{lin2018bsn,lin2019bmn,lin2020fast} in temporal action detection area, we decouple the task of video advertisement content structuring into two subtasks, i.e Temporal Segmentation and Proposal Tagging. In the temporal segmentation phrase, we apply Boundary-Matching Network (BMN) \cite{lin2019bmn} to generate temporal proposals using multi-modal representation. We then propose to refine them by scene-guided alignment and re-ranking to make the generated proposals more accurate. In the proposal tagging phrase, we take a whole video as the input and incorporate proposal located embeddings into the multi-modal encoder to capture temporal relationships between local features of each proposal and global features of the whole video. Finally, the mean pooling is adopted on the top of multi-modal representation for tagging.

\begin{figure*}[ht]
	%\vskip 0.2in
	\begin{center}		\includegraphics[width=0.85\textwidth]{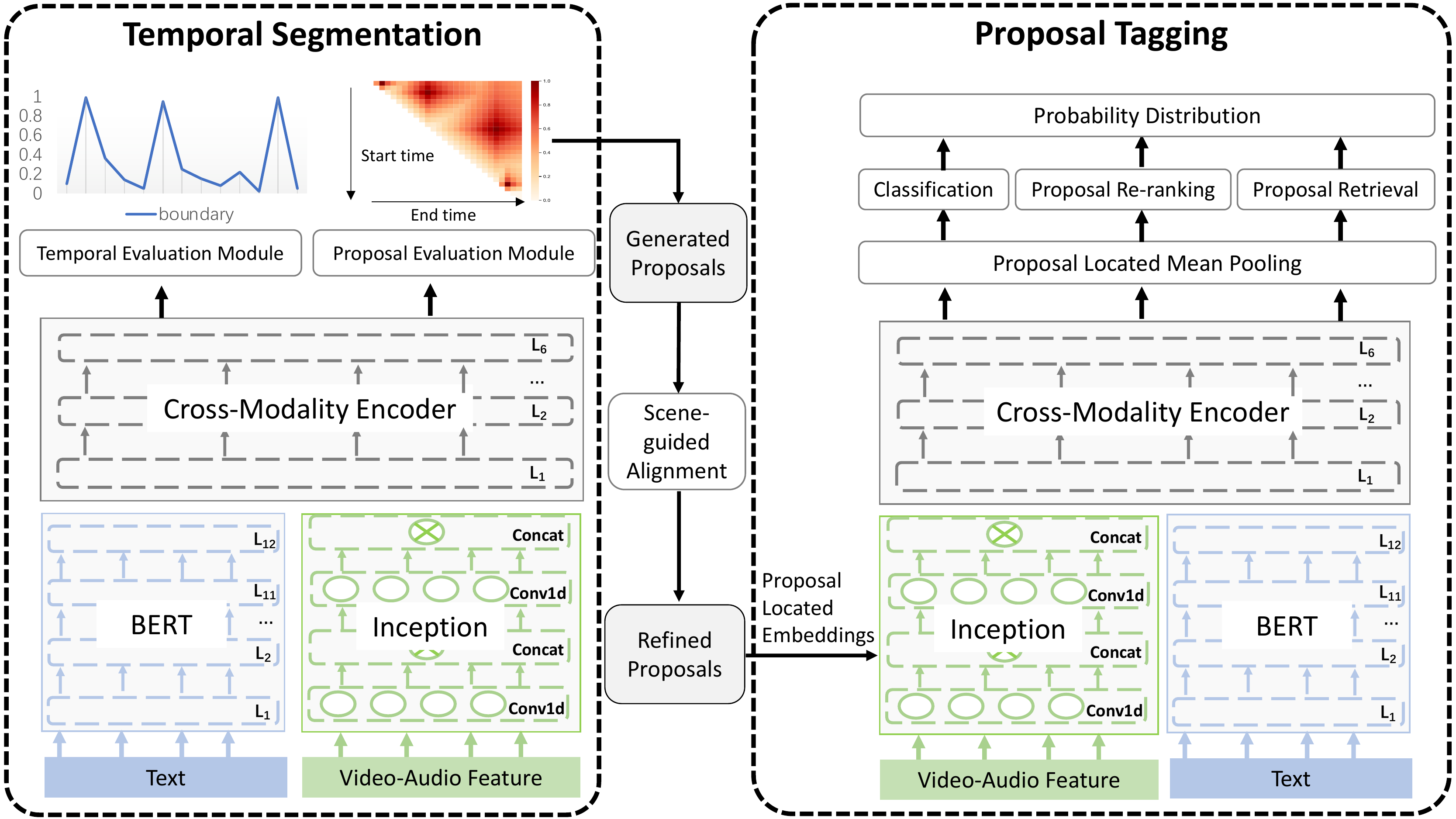}
		\caption{Overview of our proposed framework. We decouple the task of video ads content structuring into two subtasks. The first subtask is segment that generates proposals, which is shown in the left part. The second subtask is tagging that classifies each proposal, which is shown in the right part.}
		\label{figure-model}
	\end{center}
\end{figure*}

We evaluate the proposed  method on the dataset of Multi-modal Ads Video Understanding in ACM Multimedia 2021 Grand Challenge. Experiments show that our method achieves state-of-the-art performance and Rank 1 on the leaderboard. Further analysis shows that multi-modal information and newly introduced proposal located embeddings are helpful for video ads content structuring.

\section{Approach}

\subsection{Overview}
Figure \ref{figure-model} gives an overview of our approach. As shown in the Figure, we decouple the task of video ads content strutting into two subtasks, i.e. segment and tagging. We first propose
a multi-modal encoder as the backbone of two subtasks that takes text with video-audio feature as the input to obtain multi-modal representation of video advertisement.
In the segment phase, we adopt Boundary-Matching Network (BMN) \cite{lin2019bmn} to generate proposals based on the multi-modal representation. These generated proposals are further refined by scene-guided alignment \cite{pyscenedetect2017} and re-ranking, which will make proposals more accurate. In the tagging phase, we propose a proposal located embeddings (PLE) to capture temporal relation-ships between local features of each proposal and global features of the whole video for proposal tagging. In the next, we will introduce how to encode multi-modal content including video-audio and text in Section \ref{multi-modal rep}. The details about our solution on temporal segmentation and proposal tagging subtasks will be introduced in Sections \ref{segment} and Section \ref{tagging}, respectively.

\subsection{Multi-modal Encoder}
\label{multi-modal rep}
\paragraph{\textbf{Text Encoder}}
Captions in video advertisements are of significant for video ads content structuring. Taking Figure \ref{figure-case} as an example, the caption ``Hurry up and click on the bottom of the video to sign up for your child" could help the model infer this frame is a promotion page. To leverage text information in video ads, we first utilize Optical Character Recognition (OCR) technique to extract caption as the text input, denoted as $X=\{x_0,x_1,..,x_{n-1}\}$. Since pre-trained models \cite{peters2018deep,devlin2018bert} have led to strong improvement on numerous natural language processing (NLP) tasks, we use a powerful pre-trained model BERT \cite{devlin2018bert} as our text encoder to encode the text input and obtain hidden states of the text $H_X=\{h_{x_0},h_{x_1},...,h_{x_{n-1}}\}$. 

\paragraph{\textbf{Video-Audio Encoder}}
% Directly taking the whole of video and audio as the input is computationally intractable. Instead, we first split video ads into a video clip sequence. The duration of each video clip is 0.5 second. For each video clip, we sample a frame sequence and use S3D \cite{xie2018rethinking} and VGGish \cite{hershey2017cnn} to extract video and audio features, denoted as $V=\{v_0,v_1,...,v_{m-1}\}$ and $A=\{a_0,a_1,...,a_{m-1}\}$, respectively. To obtain context information from surrounding video clip, an Inception neural network \cite{szegedy2015going} implemented by one-dimensional CNN is used as the video-audio encoder to read the concatenation $Y=\{[v_0;a_0],...,[v_{m-1};a_{m-1}]\}$ of video-audio features and get context representation of video-audio $H_Y=\{h_{y_0},h_{y_1},...,h_{y_{m-1}}\}$ .

Inspired by \cite{ma2020active}, we combine video and audio features to obtain more discriminate representations. Given an input video, we first split it into video clips with length 0.5 second. For each video clip, we follow \cite{xie2018rethinking} to use S3D model pre-trained on HowTo100M dataset\cite{miech2019howto100m} to extract its visual feature. For the given audio input, we extract its feature by VGGish\cite{hershey2017cnn}. We then re-sample the video and audio feature sequence into the same temporal size $m$ using bi-linear interpolation. We denote the re-sampled video and audio features as $V=\{v_0,v_1,...,v_{m-1}\}$ and $A=\{a_0,a_1,...,a_{m-1}\}$, respectively. Since different video segment are variance in length, we follow \cite{szegedy2015going} to adopt Inception module to capture the information from different temporal sizes. Specifically, We concatenate $V$ and $A$ into $Y=\{[v_0;a_0],...,[v_{m-1};a_{m-1}]\}$ along the channel dimension to form the input video-audio features, then feed it into two Inception modules and produce $H_Y=\{h_{y_0},h_{y_1},...,h_{y_{m-1}}\}$. We follow \cite{szegedy2015going} to design the Inception module, while only replacing the 2D convolutional layers into 1D ones.

%The architecture of the Inception module can be found in Figure~\ref{}.

\paragraph{\textbf{Cross-Modality Encoder}}
The text encoder and video-audio encoder mainly focus on a part of modality. To fully leverage the text and video-audio, we adopt a 6-layer Transformer \cite{vaswani2017attention} as our cross-modality encoder for cross-modality interaction.  The input of cross-modality encoder is constructed by summing type embeddings and the concatenation $[H_X;H_Y]$ of the text and video-audio representation. Finally, we obtain multi-modal representation $H'=\{h'_{x_0},...,h'_{x_{n-1}},h'_{v_0},...,h'_{v_{m-1}}\}$. We denote the multi-modal representation of the whole video ads as $H=\{h'_{v_0},h'_{v_1},...,h'_{v_{m-1}}\}$, which will be used for segment and tagging phases.

\subsection{Temporal Segmentation}
\paragraph{\textbf{Boundary-Matching Network}}
\label{segment}
In the task of temporal action detection, \citet{lin2019bmn} propose the Boundary-Matching Network (BMN) to generate high-quality temporal proposals. The network consists of two components, including Temporal Evaluation Module (TEM) and Proposal Evaluation Module (PEM). TEM aims to predict precise boundaries for all temporal locations in untrimmed video and PEM aims to provide confidence for each proposal.

Different from datasets like ActivityNet \cite{caba2015activitynet} and THUMOS \cite{idrees2017thumos}, there are no backgrounds in video ads content structuring. Therefore, TEM only predicts boundary probability of two segments for all temporal locations, shown in the top left of segment part in Figure \ref{figure-model}. 
We use 3-layer CNN following by a 6-layer Transformer as the TEM and take multi-modal representation $H$ as the input to calculate boundary probability of each video clip $P^{Tem}=\{p^{Tem}_0,..,p^{Tem}_{m-1}\}$. We use the cross entropy (CE) loss to train TEM, where $y_i\in\{0,1\}$ is the boundary label for $i$-th video clip.
\begin{equation}
loss_{TEM}=-\frac{1}{m}\sum_{i=0}^{m-1}[y_ilogp^{TEM}_i+(1-y_i)log(1-p^{TEM}_i)]
\end{equation}

For proposal evaluation module shown in the top right of segment part in Figure \ref{figure-model}, we use the same network architecture and loss function $loss_{PEM}$ as \citet{lin2019bmn} and provide a confidence $p^{cof}_{ij}$ for each proposal $prop_{ij}$ from $i$-th to $j$-th video clip. The final loss function of Boundary-Matching Network is empirically as:
\begin{equation}
loss_{BMN}=5 \cdot loss_{TEM}+loss_{PEM}
\end{equation}

%The network architecture and loss function we use in PEM is the same as \citet{lin2019bmn}. To generate more reliable confidence scores, for each proposal from $i$-th to $j$-th video clip, we fuse its cut-point probabilities and confidence scores by multiplication to generate the final confidence score $s_{ij}$:
%\begin{equation}
%s_{ij} = p_i*p_j*s^{cof}_{ij}*[(1-p_{i+1})*(1-p_{i+2})*\cdots*(1-p_{j-1})]
%\end{equation}
%where $(1-p_{i+1})*(1-p_{i+2})*\cdots*(1-p_{j-1})$ is the probability of no existing cut- point between $i$-th and $j$-th video clip. Since the number of video clips for each proposal is different, we approximate the confidence score as the following equation:
%\begin{equation}
%s_{ij} = p_i*p_j*s^{cof}_{ij}*min\{(1-p_{i+1}),(1-p_{i+2}),...,(1-p_{j-1})\}
%\end{equation}

In the inference phase of BMN, for each proposal $prop_{ij}$, we fuse its boundary probabilities and confidence scores by multiplication to generate the final confidence score $s_{ij}$:
\begin{equation}
\label{score_proposal}
p^{prop}_{ij} = p_i \cdot p_j \cdot p^{cof}_{ij} \cdot min\{\overline{p_{i+1}},\overline{p_{i+2}},...,\overline{p_{j-1}}\} 
\end{equation}
where $\overline{p_k}=1-p_k$. Different from the final confidence score of \citet{lin2019bmn}, we add the last term to indicate the approximate probability of no existing boundary in the middle of the proposal $prop_{ij}$. The main reason for using approximate probability is that true probability will over-punish proposals that too long.

Finally, we use Non-Maximum Suppression (NMS) algorithm to generate non-overlap proposal by the confidence score $s_{ij}$. 

%Limited by the amount of calculation and memory, we only set the duration of each video clip as 0.5 second. 

\paragraph{\textbf{Scene-Guided Alignment}}
The duration of each video clip we use for temporal segmentation is only 0.5 second, thus the temporal boundaries predicted by BMN are rough. To make the predicted boundaries more precise, we propose to use the scene-changed frames to fine-adjust the predicted boundaries. We call this method scene-guided alignment (SGA). We follow \cite{pyscenedetect2017} to extract all frames with scene probability greater than 0.1 for each given video. For each predicted boundary, if the temporal error between it and its nearest scene frame is less than 0.5 second, we will move the boundary to the position of its nearest scene frame.

% To further refine generated proposal and make them more accurate, we refine them by scene-guided alignment algorithm \cite{pyscenedetect2017}.

\subsection{Proposal Tagging}
\label{tagging}

\paragraph{\textbf{Proposal Located Embedding}}
%We choose multi-modal encoder introduced in Section \ref{multi-modal rep} as the backbone of our classification model. 
To incorporate proposal temporal information for proposal tagging, we introduce proposal located embedding before the Inception modules. Specially, given a proposal $prop_{ij}$, the input of Inception module changes from $Y$ to $Y^t=\{[v_0;a_0]+t_0,...,[v_{m-1};a_{m-1}]+t_{m-1}\}$, where $t_k$ is a trainable randomly initialized embedding to indicate proposal location if $i \leq k \leq j$ otherwise $t_k$ is another trainable randomly initialized embedding to indicate non-proposal location. After obtaining multi-modal representation of video-audio from multi-modal encoder, we get the final vector $v_{ij}$ by mean pooling over the location of the proposal $prop_{ij}$. Finally, we leverage the final vector for proposal tagging in a multi-task manner.

\paragraph{\textbf{Classification}}
The proposal tagging task can be formulate as a multi-label classification problem. We use a fully connected layer with sigmoid activation for classification, and adopt binary cross entropy as loss function. The probabilities of classification for the proposal $prop_{ij}$ are denoted as $p^{cls}_{ij}$.

%In the inference phase, the probability distribution of the proposal from $i$-th to $j$-th video clip is calculated as $P^C_{ij}*P^R_{ij}$. 

\paragraph{\textbf{Proposal Re-ranking}}
To make proposal more accurate, we introduce a re-ranking task in the classifier. The task aims to re-rank generated proposals, which will improve the precision of generated proposals. Specially, we use a fully connected layers with sigmoid activation to predict Intersection over Union (IoU) score for each generated proposal $prop_{ij}$, denote as $p^{iou}_{ij}$. 

%In the inference phase, the probability distribution of the proposal from $i$-th to $j$-th video clipis is calculated as $P^C_{ij}*P^R_{ij}*s^{iou}_{ij}$. 

\paragraph{\textbf{Proposal Retrieval}}
% Different from general videos, we find that there is a lot of similarity between segments, such as presentation form, scene, and style.
For video ads content structuring task, it would have more application value if it can be salable to some new scenes or categories. Inspired by \citet{yang2021garbagenet}, we propose to utilize multi-modal representation $v_{ij}$ of the proposal $prop_{ij}$ to retrieve top 10 most similar segments using cosine similarity. We denote labels and similarity score of these retrieved segments as $\{g^0_{ij},g^1_{ij},...,g^9_{ij}\}$ and $\{c^0_{ij},c^1_{ij},...,c^9_{ij}\}$, respectively. The retrieved result is calculated by weighted summing retrieved labels:
\begin{equation}
\label{retrieval}
p^{ret}_{ij}=\frac{\sum_{k=0}^{9}c^k_{ij}g^k_{ij}}{\sum_{k=0}^{9}c^k_{ij}}
\end{equation}

In the inference phase, the probability distribution $p^{cat}_{ij}$ of categories of the proposal $prop_{ij}$ is calculated as:
\begin{equation}
p^{cat}_{ij}= p^{*}_{ij}\cdot p^{iou}_{ij} \cdot p^{prop}_{ij},
\end{equation}
where $p^{*}_{ij}$ can be $p^{cls}_{ij}$, $p^{ret}_{ij}$, or the combination of them.

\section{Experiments}
\subsection{Experiments Setup}

\paragraph{\textbf{Dataset.}} We evaluate our proposed approach on  the dataset of Multi-modal Ads Video Understanding in ACM Multimedia 2021 Grand Challenge. The dataset consists of $5,000$ videos. Each video is split into one or several clips by annotators. Each video clip is annotated by at least one categories. The total number of category is $82$. When performing evaluation, 
the task constraints that predicted proposals can not have overlap with each other. To avoid redundancy prediction, the task also limits that a proposal only produce 20 category labels.

\paragraph{\textbf{Training Detail.}} All our experiments are conducted on one NVIDIA Tesla V100-32G GPU. We use AdamW optimizer with 1e-4 learning rate to train all models for 10 epochs and evaluate the model using 5-fold cross-validation.

\paragraph{\textbf{Evaluation.}} 
The goal of temporal segmentation is to generate high quality proposals to cover ground truth segments with high recall and high temporal overlap. To evaluate proposal quality, we follow \citet{lin2019bmn} to use AUC under IoU thresholds [0.5 : 0.05 : 0.95] as a metric. 
Beyond that, we also  use F1-score between predicted and ground truth boundaries as another metric to evaluate the precision of proposals. Given a prediction of a video, if a predicted boundary can match any ground truth boundaries within 0.5s error, it will be consider as a true positive prediction, and otherwise will be consider as a false positive prediction. Note that one ground truth boundary will be only matched once. 
Finally, the overall performance of generated proposals is the product of AUC and F1-score. 
To evaluate the performance of proposal tagging, we follow \citet{caba2015activitynet} to use mAP@[0.5:0.05:0.95] as the metric.
%We use F1-score and AUC as the metric to evaluate the quality of video proposals. Given a prediction of a video, if a predicted boundary can match any ground truth boundaries within 0.5s error, it will be consider as a true positive prediction, and otherwise will be consider as a false positive prediction. Note that one ground truth boundary will be only matched once. We then calculate F1-score between predicted and ground truth  boundaries as the metric. To evaluate the performance of proposal tagging, We follow \citet{caba2015activitynet} to use mAP@[0.5:0.05:0.95] as the metric.

\subsection{Evaluation on Temporal Segmentation}

\begin{table}[h]
\small
    \centering
    \begin{tabular}{c|c|c|c|c|c|c}
\hline
\textbf{Model} & \textbf{Video} & \textbf{Audio} & \textbf{Text} & \textbf{AUC} & \textbf{F1} & \textbf{Overall}\\
\hline
BMN \cite{lin2019bmn} & \checkmark & & & 72.1 & 78.7 &56.7\\
\hline
\multirow{3}{*}{Ours} & \checkmark & &  & 74.8& 79.0&59.1 \\
& \checkmark & \checkmark & & \bf{75.1}& 78.6&59.0  \\
& \bf{\checkmark} & \bf{\checkmark} & \bf{\checkmark} & 74.4& \bf{80.9}&\bf{60.2} \\
\hline
    \end{tabular}
    \caption{Results on video temporal segmentation task.}
    \label{tab:segment_ablation}
    \vspace{-0.2in}
\end{table}

We take the state-of-the-art model in temporal action detection as our baseline, i.e. BNN \cite{lin2019bmn}. The difference between our segment model and the baseline includes: (1) We add the last term to indicate the approximate probability of no existing boundary in the middle of the proposal in Equation \ref{score_proposal}. (2) we leverage multi-modal content. 

Table \ref{tab:segment_ablation} show experiment results on temporal segmentation task. We can see that our approach achieve a 3.5 gain of overall score, which significantly outperforms the baseline. From Table~\ref{tab:segment_ablation}, we find that incorporating the audio feature improves the AUC score by 0.3 but will hurt the F1-score, which shows that audio feature may not have much effect on temporal segmentation task. After leveraging text information, we can see that the interaction with text bring 1.1 gain of overall score compared with the model that only leverages video, which reveals the importance of multi-modal representation.

\subsection{Evaluation on Proposal Tagging}

\begin{table}[h]
\small
    \centering
    \begin{tabular}{c|c|c|c|c}
\hline
\textbf{Model} & \textbf{Video} & \textbf{Audio} & \textbf{Text} & \textbf{mAP}\\
\hline
Inception \cite{szegedy2015going} w/o PLE& \checkmark & & & 27.3 \\
\hline
\multirow{3}{*}{Ours} & \checkmark & & &  27.9 \\
& \checkmark & \checkmark & & 28.1 \\
& \checkmark & \checkmark & \checkmark & \bf{29.5} \\
\hline
    \end{tabular}
    \caption{Results on proposal tagging task.}
    \label{tab:tagging_ablation_1}
\vspace{-0.2in}
\end{table}
We report the experiment results on proposal tagging task in Table~\ref{tab:tagging_ablation_1}. In this experiment, we take Inception \cite{szegedy2015going} without proposal located embedding ({\bf Inception w/o PLE}) as a baseline, which only use video feature and remove PLE in classification module. For fair comparison, all settings in Table \ref{tab:tagging_ablation_1} use the same generated proposals generated by best model in Table \ref{tab:segment_ablation}.

When only leveraging the same video feature in Table \ref{tab:tagging_ablation_1}, we can see that incorporating PLE into the Inception bring 0.6 gain of mAP score, which demonstrates that the proposal located embedding could help proposal tagging. After levering multi-modal content like speech and caption, results show that our multi-modal encoder significantly outperforms the model that only uses single-modal, which shows the effectiveness of our multi-modal representation.

\subsection{Classification-Vs. Retrieval-based Classifier}

\begin{table}[h]
\small
    \centering
    \begin{tabular}{c|c}
\hline
\textbf{Classifier} &  \textbf{mAP}\\
\hline
Classification-based method & 29.8 \\
Retrieval-based method & 30.3  \\
Ensemble method &  \textbf{31.7} \\
\hline
    \end{tabular}
    \caption{Results of various classifiers.}
    \label{tab:tagging_ablation_2}
    \vspace{-0.2in}
\end{table}

In real application scenario, the category number may be updated frequently. Retrieval-based method that retrieves similar examples from training dataset and infers probability distribution of categories of the proposal may bring more application value. In the Table~\ref{tab:tagging_ablation_2}, we show the performance of various classifiers. The probability distribution of categories of the proposal $prop_{ij}$ in classification-based method is calculated as $p^{cls}_{ij}$ described in Section \ref{tagging}, while the probability distribution $p^{ret}_{ij}$ in retrieval-based method is calculated by retrieving similar proposal as Equation \ref{retrieval}.

We can find that, the retrieval-based method can achieve comparable result with the classification-based method. To achieve higher evaluation score, we ensemble the results from both classification-based and retrieval-based methods, and find that the ensemble model brings further improvements with 1.4\% absolute gain. 

%We also train another model based on only $40$ categories, and we can find that the retrieval-based method also can achieve xxx mAP, which only slightly lower than using full training data.

\section{Conclusion}
In this paper, we propose a multi-modal encoder to learn multi-modal representation from video advertisements by interacting between video-audio and text. Experiments show that multi-modal representation significantly improve temporal segmentation and proposal tagging tasks. Based on multi-modal representation, we present an efficient framework for the task of video ads content structuring. The framework achieves Rank 1 on the task of Multi-modal Ads Video Understanding in ACM Multimedia 2021 Grand Challenge. In future work, we would like to explore how to pre-train a powerful multi-modal encoder using video ads for empowering video ads content analysis.

\bibliographystyle{ACM-Reference-Format}
\bibliography{sample-base}
\end{document}